%% file: eccv2020submission.tex
\pdfoutput=1
\documentclass[runningheads]{llncs}
\usepackage{graphicx}

\usepackage{tikz}
\usepackage{comment}
\usepackage{amsmath,amssymb} 
\usepackage{color}
\usepackage{cite}
\usepackage{multirow}
\usepackage{booktabs}
\usepackage{xcolor}
\usepackage{url}

\begin{document}
\pagestyle{headings}
\mainmatter
\def\ECCVSubNumber{9}  

\title{A Deep Dive into Adversarial Robustness in Zero-Shot Learning} 

\titlerunning{Adversarial Robustness in Zero-Shot Learning}
%
\author{Mehmet Kerim Yucel\inst{1}\and
Ramazan Gokberk Cinbis \inst{2}\and
Pinar Duygulu \inst{1}}
\authorrunning{Yucel et al.}
%
\institute{Hacettepe University, Department of Computer Engineering \and
Middle East Technical University (METU), Department of Computer Engineering}
\maketitle
\setcounter{footnote}{0}
\begin{abstract}
\input{abstract3.tex}
\end{abstract}
\input{introduction2.tex}

\input{related_work}
\input{method}

\input{exp}
\input{conclusion}

{\small
\bibliographystyle{splncs04}
\bibliography{egbib}
}

\end{document}

%% file: abstract3.tex
\textcolor{black}{Machine learning (ML) systems have introduced significant advances in various fields, due to the introduction of highly complex models}. Despite their success, it has been shown multiple times that machine learning models are prone to imperceptible perturbations that can severely degrade their accuracy. So far, existing studies have primarily focused on models where supervision across all classes were  available. In constrast, Zero-shot Learning (ZSL) and Generalized Zero-shot Learning (GZSL) tasks inherently lack supervision across all classes. In this paper, we present a study aimed on evaluating the adversarial robustness of ZSL and GZSL models. We leverage the well-established label embedding model and subject it to  a set of established adversarial attacks and defenses across multiple datasets. In addition to creating possibly the first benchmark on adversarial robustness of ZSL models, we also present analyses on important points that require attention for better interpretation of ZSL robustness results. We hope these points, along with the benchmark, will help researchers establish a better understanding what challenges lie ahead and help guide their work.

%% file: introduction2.tex
\section{Introduction}
\label{sec:intro}

The meteoric rise of complex machine learning models in the last decade sparked a whole new wave of state-of-the-art (SOTA) results in numerous fields, such as computer vision, low-level image processing, natural language processing and speech recognition. Due to the increase in available data, compute power and architectural improvements, these fields are still seeing rapid improvements in various tasks with no signs of slowing down.

However, it has been shown \cite{szegedy2013intriguing} that ML models are prone to adversarial examples, which are perturbations aimed to guide models into inaccurate results. Such perturbations can successfully misguide models while introducing imperceptible perturbations to a query data. Starting with computer vision, such attacks have been extended to speech recognition \cite{carlini2018audio}, natural language processing \cite{ebrahimi2017hotflip} and various other tasks/modalities \cite{cao2019adversarial}. Naturally, equal attention has been given to defend the models against these attacks, either by designing robust models or introducing mechanisms to detect and invalidate adversarial examples \cite{papernot2016distillation}. Adversarial machine learning initially focused on small datasets in computer vision, such as MNIST and CIFAR-10, but it has been extended to large datasets such as ImageNet and even commercial products \cite{yuan2019adversarial, papernot2017practical}. 

The majority of the adversarial ML literature has so far focused on supervised models and aimed to improve their robustness using various approaches. In Zero-shot learning (ZSL) and Generalized Zero-shot learning (GZSL) settings, however, the task differs from a generic supervised approach; the aim is to learn from a set of classes such that we can optimize the knowledge transfer from these classes to a set of previously unseen classes, which we use during the evaluation phase. As a result, the notorious problem of ZSL is far from being solved despite the significant advances made in the last few years. The introduction of adversarial examples to ZSL models would theoretically further exacerbate the problem.

In this paper, we present an exploratory adversarial robustness analysis on ZSL techniques \footnote{Code is available at \url{https://github.com/MKYucel/adversarial_robustness_zsl}
}. Unlike the most recent approaches where ZSL problem is effectively reduced to a supervised problem \cite{sariyildiz2019gradient, bucher2017generating}, we take a a step back and focus on label embedding model \textcolor{black}{\cite{akata2013label, weston2010large}} and analyze its robustness against several prominent adversarial attacks and defenses. Through rigorous evaluation on most widely used ZSL datasets, we establish a framework where we analyse not only the algorithm itself, but also the effect of each dataset, effect of per-class sample count, the trends in boundary transitions as well as how the existing knowledge transfer between seen and unseen classes are effected by adversarial intrusions.  We hope the presented framework will focus the community's attention on the robustness of ZSL models, which has largely been neglected. Moreover, this study will serve as a benchmark for future studies and shed light on important trends to look out for when analysing ZSL models for adversarial robustness.

%% file: related_work.tex
\section{Related Work} \label{related_work}

\textbf{Adversarial attacks.} Adversarial ML has been an integral part of ML research in the last few years as it exposed significant robustness-related problems with existing models. It has first been shown in \cite{szegedy2013intriguing} that a perturbation can be crafted by optimizing for misclassification of an image; under certain $\ell$-norm constraints this perturbation can even be imperceptible to humans. A fast, one-step attack that exploits the gradients of the cost function w.r.t the model parameters to craft a perturbation was shown in \cite{goodfellow2014explaining}. An iterative algorithm that approximates the minimum perturbation required for misclassification in ML models has been shown in \cite{moosavi2016deepfool}.  Carlini and Wagner \cite{carlini2017towards} showed that a much improved version of \cite{szegedy2013intriguing} can be tailored into three different attacks, each using a different $\ell$-norm, that can succesfully scale to ImageNet and invalidate the distillation defense \cite{papernot2016distillation}. 

Transformation attacks \cite{xiao2018spatially}, universal attacks for a dataset \cite{moosavi2017universal}, one-pixel attacks \cite{su2019one}, attacks focusing on finding the pixels that will alter the output the most \cite{papernot2016limitations},  unrestricted attacks \cite{bhattad2020unrestricted}, black-box attacks where attacker has no information about the model \cite{papernot2017practical} and attacks transferred to the physical world \cite{kurakin2016adversarial} are some highlights of adversarial ML. Moreover, these attacks have been extended to various other ML models with various modalities, such as recurrent models \cite{hu2018black}, reinforcement learning \cite{behzadan2017vulnerability}, object detection \cite{xie2017adversarial}, tracking \cite{chen2020one, jia2019fooling}, natural language processing \cite{ebrahimi2017hotflip}, semantic segmentation \cite{arnab2018robustness}, graph neural networks \cite{zugner2019adversarial}, networks trained on LIDAR data \cite{cao2019adversarial} , speech recognition \cite{carlini2018audio}, visual question answering \cite{sharma2018attend} and even commercial systems \cite{liu2016delving}.

\textbf{Adversarial defenses.} In order to address the robustness concerns raised by adversarial attacks, various defense mechanisms have been proposed, resulting into an arms-race in the literature between attacks and defenses. Several prominent defense techniques include network distillation \cite{papernot2016distillation}, adversarial training \cite{szegedy2013intriguing}, label smoothing \cite{hazan2016perturbations}, input-gradient regularization \cite{ross2018improving}, analysis of ReLu activation patterns \cite{lu2017safetynet}, feature regeneration \cite{Borkar_2020_CVPR}, generalizable defenses \cite{naseer2020self},  exploitation of GANs \cite{samangouei2018defensegan} and auto-encoders \cite{meng2017magnet} for alleviation of adversarial changes and feature space regularization via metric learning \cite{mao2019metric}. These defense methods rely on either re-training of the network or preparing an additional module that would either detect or alleviate the adversarial perturbations. Another segment of adversarial defenses have borrowed several existing techniques and used them towards adversarial robustness; JPEG compression \cite{shaham2018defending}, bit-depth reduction, spatial smoothing \cite{xu2017squeeze} and total variance minimization \cite{guo2018countering} are some examples of such techniques. Adversarial ML field is quite vast; readers are referred to \cite{survey2018adversarial, Xu2019AdversarialAA} for a more in-depth discussion.

\textbf{Zero-shot learning}. In majority of the ML fields, SOTA results are generally held by supervised models, where all classes have a form of strong supervisory signal (i.e. ground-truth labels) that guides the learning. \textcolor{black}{However, the collection of a supervised training set quickly turns into a bottleneck in semantically scaling up a recognition model}. The problem of strong supervision through ground-truth labels can be somewhat alleviated by the (transductive) self-supervised learning \cite{selfsupervised2020survey}, but the unlabeled data or the auxiliary supervision may not be available for every task.  Zero-shot learning aims to address this issue by bridging the gap between \textit{seen} (i.e. classes available during training) and \textit{unseen} (i.e. classes unavailable during training) classes by transfering the knowledge learned during training. \textcolor{black}{Generalized Zero-Shot learning, on the other hand, aims to facilitate this knowledge transfer while keeping the accuracy levels on the \textit{seen} classes as high as possible}. The auxiliary information present for both seen and unseen classes (i.e. class embeddings) is exploited to bridge the gap. 

In its early years, ZSL methods consisted of a two-stage mechanism, where the attributes of an image were predicted and these attributes were used to find the class with the most similar attributes \cite{dap2013zsl, twostage2009zsl}. By directly learning a linear \cite{akata2013label, devise2013nips, Akata_2015_CVPR, Li_2018_CVPR} or a non-linear compability \cite{Xian_2016_CVPR, crossmodal2013nips, Chen_2018_CVPR, semantic2018nips} function to map from visual space to a semantic space, later models transitioned to a single-stage format. The reverse mapping, from semantic space to the visual space \cite{Wang_2018_CVPR, Zhang_2017_CVPR}, has also been explored for ZSL problem. Embedding both visual and semantic embeddings into a common latent space for ZSL have also proven to be successfull \cite{Zhang_2015_ICCV, Changpinyo_2016_CVPR}. Transductive approaches leveraging visual or semantic information on unlabeled unseen classes \cite{Ye_2017_CVPR, trans2017icmr} are considered another discriminative approach for ZSL. In recent years, \textcolor{black}{in addition to discriminative approaches \cite{jiang2019transferable, zhang2018triple}}, generative approaches  \cite{sariyildiz2019gradient, bucher2017generating, kumar2018generalized, felix2018multi} which model the mapping between visual and semantic spaces are increasingly being used to generate samples for unseen classes, slowly reducing ZSL to a fully-supervised problem. For further information on ZSL, readers are referred to \cite{gbu2018pami,  zsl2019survey}.

\textcolor{black}{Only a recent unpublished study} \cite{Zhang2019ATZSLDZ} proposed a ZSL model that is robust to several adversarial attacks by formulating an adversarial training regime. Our study, on the other hand, concentrates on setting up a framework and creates a benchmark to guide researchers' efforts towards adversarially robust ZSL/GZSL models, by presenting a detailed analysis of existing datasets and the effects of several well-established attacks and defenses. \textcolor{black}{To the best of our knowledge, our study is the first to establish such a benchmark with a detailed analysis.}

%% file: method.tex
\section{Methodology} 

\subsection{Model Selection}
For the model selection, we take a step back and focus on the models that aim to transfer the knowledge learned from seen classes to unseen classes, unlike the recent studies that rely on generative models to generate samples using class embeddings for unseen classes and try to reduce ZSL to a fully supervised setting. We hypothesize that concentrating on the latter would mean evaluating the sample generation part for adversarial robustness, rather than evaluating  the robustness of the model that aims directly to facilitate seen/unseen class knowledge transfer.

As presented in Section \ref{related_work}, there are numerous suitable candidates for our goal. Towards this end, we select the label-embedding model \cite{akata2013label}, \textcolor{black}{which has been shown to be a stable and competitive model in modern benchmarks \cite{gbu2018pami}}. Attribute-label embedding (ALE) model is formulated as

\begin{equation}
F(x,y;W) = \theta (x)W^{T}\phi (y) 
\end{equation}
where $\theta(x)$ is the visual and $\phi(y)$ is the class embeddings. These two modalities are associated \textcolor{black}{through the compatibility function} $F()$, which is parametrized by learnable weights \textit{W}.

The reason for selecting ALE can be described by the fact that it is one of the earlier studies that showed direct mapping by exploiting data and auxiliary information is more effective than intermediate attribute prediction stages. Although there are several studies which build on what ALE does \footnote{As noted in \cite{gbu2018pami}, models focusing on linear compability functions have the same formulation, but different optimization objectives.}, we believe  results of ALE will be representative of the adversarial robustness of this family of ZSL approaches. Individual analyses of more approaches are certainly welcome, but is not in our scope. It must be noted that we purely focus on an inductive setting for ZSL.

\subsection{Attacks and Defenses}

\textbf{Threat Model.} Our evaluation makes several assumptions on the threat model. We choose three white-box, per-image attacks (i.e. non-universal) where the attacker has access to model architecture and its parameters. The attack model operates under a setting where \textit{all} images are attacked, regardless of their original predictions (i.e. whether they were classified correctly by the model or not).
We choose a training-time defense (i.e. robustifying the model by re-training) and two data-independent, pre-processing defenses, where input images are processed before being fed to the network.  The defense model operates under a \textit{blind} setting, where none of the defenders have access to attack details or the attack frequency (i.e. defenses are applied to all images; regardless of the fact that attacks introduced misclassifications or not). \textcolor{black}{In the next sections, we briefly present attacks and defenses considered in this work. We do not provide details due to page limitations}.

\textbf{Attacks.} \textit{The first attack} we select is the widely-used Fast Gradient Sign method (FGSM) attack \cite{goodfellow2014explaining} that is based on the \textit{linearity hypothesis}. By taking the gradient of the loss function with respect to the input, change of the output with respect to each input component is effectively estimated. This information is then used to craft adversarial perturbations that will guide the image towards these directions, which means \textit{maximizing} the loss with respect to input components. We select FGSM due to its one-shot nature (i.e. no optimization), its low computational complexity and the fact that it is inherently \textit{not} optimized for the minimum possible perturbation.

\textit{The second attack} is the \textit{DeepFool} \cite{moosavi2016deepfool} attack. DeepFool concentrates on the distance of an image to the closest decision boundary. Essentially, DeepFool calculates the distance to select number of decision boundaries, finds the closest one and takes the step towards this boundary. For non-linear classifiers, this step is approximated by an iterative regime that repeatedly tries to cross the boundary, until an iteration limit is reached or the boundary is crossed. We select DeepFool due to several reasons; i) it is an optimization based attack, ii) it directly aims for the minimum perturbation, iii) it operates under the linearity hypothesis assumption and iv) it is inherently indicative of the decision boundary characteristics. The version of DeepFool we experiment with is the original \textit{untargeted} version that controls the perturbation with the $\ell_2$ norm.

\textit{The last attack } is the \textit{Carlini-Wagner} \cite{carlini2017towards} attack. In their paper, authors essentially refine the objective function proposed in \cite{szegedy2013intriguing} via several improvements and propose three different attacks where each attack uses a different $\ell$-norm constraint to control the perturbation. We select Carlini-Wagner attacks due to several reasons; i) it is one of the first attacks that is shown to beat an adversarial defense, ii) one of the first to scale to ImageNet and iii) it is still one of the high-performing attacks in the literature.  \textcolor{black}{We use the \textit{untargeted}, $\ell_2$-norm version to have a better comparison with DeepFool}.

\hfill \\ 
\hfill \\ 

\textbf{Defenses.} \textit{The first defense} analyzed is the well-known \textit{label smoothing}. It is a well-known  regularization technique that prevents over-confident predictions. Label smoothing has been shown to be a good defense mechanism  \cite{hazan2016perturbations, labelsmoothadv2020} and its success is tied to the prevention of confident classifier predictions when faced with an out-of-distribution sample. We select label smoothing as it is i) a training-time defense, ii) conceptually easy and iii) it is a good use case of ZSL models.

\textit{The second defense} is \textit{local spatial smoothing}. It has been reported that feature squeezing techniques \cite{xu2017squeeze} can provide robustness against adversarial examples as they effectively shrink the feature space where adversarial examples can reside. Similar to the original paper, we use median-filter with reflect-padding to pre-process images before they are fed to the model. We select spatial smoothing due to i) its data and attack-independent and ii) its inexpensive nature. Moreover, testing this against non-$l_0$ attacks is a good use case for its efficiency \footnote{It has been noted in \cite{xu2017squeeze} that this defense is inherently more effective against $l_0$-norm attacks. }. We do not use the detection mechanism in the original paper, but just the spatial smoothing operation.

\textit{The last defense} is the \textit{total variance minimization} defense. It has been proposed \cite{guo2018countering} as an input transformation defense, where the aim is to remove perturbations by image reconstruction. Initially, several pixels are selected with a Bernoulli random variable  from the perturbed image. Using the selected pixels, the image is reconstructed by taking into account the total variation measure. Total-variance minimization is shown to be an efficient defense as it encourages the removal of small and localized perturbations. We select this defense due to its simple and data/attack independent nature. It is also a good candidate to evaluate different attacks due to its localized perturbation removal ability.

%% file: exp.tex
\section{Experimental Results} \label{exp_results}
\subsection{Dataset and Evaluation Metrics} \label{exp}

\begingroup
\setlength{\tabcolsep}{8pt} 
\renewcommand{\arraystretch}{1.5} 
\begin{table}[]
\resizebox{\textwidth}{!}{%
\begin{tabular}{cccc|ccccccccc}
\multicolumn{1}{l}{}       & \multicolumn{3}{c|}{\textbf{Zero Shot}} & \multicolumn{9}{c}{\textbf{Generalized Zero Shot}}                                                        \\
\multicolumn{1}{l}{}       & C           & S           & A           & \multicolumn{3}{c|}{C}                  & \multicolumn{3}{c|}{S}                  & \multicolumn{3}{c}{A} \\ \hline
\multicolumn{1}{l}{Attack} & \multicolumn{3}{c|}{Top-1}              & u    & s    & \multicolumn{1}{c|}{h}    & u    & s    & \multicolumn{1}{c|}{h}    & u     & s     & h     \\ \hline
Original                   & 54.5        & 57.4        & 62.0        & 25.6 & 64.6 & \multicolumn{1}{c|}{36.7} & 20.5 & 32.3 & \multicolumn{1}{c|}{25.1} & 15.3  & 78.8  & 25.7  \\ \hline
$FGSM_1$                   & 40.3        & 47.7        & 42.5        & 18.5 & 45.4 & \multicolumn{1}{c|}{26.3} & 17.7 & 25.9 & \multicolumn{1}{c|}{21.0} & 10.7  & 58.9  & 18.1  \\
$FGSM_2$                   & 18.5        & 16.3        & 14.8        & 10.8 & 11.7 & \multicolumn{1}{c|}{11.2} & 8.1 & 9.8 & \multicolumn{1}{c|}{8.9} & 3.4  & 10.0  & 5.1  \\
$FGSM_3$                   & 15.2        & 11.8        & 16.4        & 9.0 & 10.2 & \multicolumn{1}{c|}{9.6} & 4.3 & 5.5 & \multicolumn{1}{c|}{4.9} & 2.2  & 11.2  & 3.7  \\ \hline
$DEFO_1$                   & 30.9        & 25.6        & 50.6        & 9.1 & 19.1 & \multicolumn{1}{c|}{12.3} & 6.4 & 7.2 & \multicolumn{1}{c|}{6.8} & 13.3  & 41.2  & 20.1  \\
$DEFO_2$                   & 30.8        & 25.5        & 50.5        & 9.1 & 18.9 & \multicolumn{1}{c|}{12.3} & 6.4 & 7.2 & \multicolumn{1}{c|}{6.8} & 13.4  & 41.2  & 20.2  \\
$DEFO_3$                   & 22.4        & 17.8        & 41.4        & 7.6 & 11.5 & \multicolumn{1}{c|}{9.2} & 6.3 & 6.3 & \multicolumn{1}{c|}{6.3} & 13.0  & 30.2  & 18.2  \\ \hline
$CaWa_{1}$                 & 28.9        & 43.1        & 43.2        & 17.0 & 29.0 & \multicolumn{1}{c|}{21.4} & 17.7 & 24.9 & \multicolumn{1}{c|}{20.7} & 15.2  & 56.3  & 24.0  \\
$CaWa_{2}$                 & 25.9        & 40.9        & 36.9        & 16.4 & 24.4 & \multicolumn{1}{c|}{19.6} & 17.7 & 23.9 & \multicolumn{1}{c|}{20.3} & 15.2  & 46.6  & 22.9  \\
$CaWa_{3}$                 & 24.6        & 39.8        & 34.7        & 15.9 & 23.1 & \multicolumn{1}{c|}{18.9} & 17.5 & 23.4 & \multicolumn{1}{c|}{20.0} & 15.2  & 43.6  & 22.5 
\end{tabular}%
}
\caption{Results when \textit{all} images are attacked. \textit{C}, \textit{S} and \textit{A} stand for CUB, SUN and AWA2 datasets, respectively. Parameters: [$FGSM_{1-3}$ $\epsilon$: 0.001, 0.01, 0.1] [$DeepFool_{1-3}$ $max\_iter, \epsilon$: (3,1e-6), (3,1e-5), (10,1e-6)] [$C\&W_{1-3}$ $max\_iter$: 3,6,10 ]. \textit{Top-1} is the top-1 accuracy, where \textit{u}, \textit{s} and \textit{h} are unseen, seen and harmonic accuracy values, respectively. }
\label{tab:attacks}
    \vspace{-8mm}
\end{table}

\endgroup

We perform our evaluation on three widely used ZSL/GZSL datasets; Caltech-UCSD-Birds 200-2011 (CUB) \cite{cubdataset}, Animals with Attributes 2 (AWA2) \cite{gbu2018pami} and SUN \cite{patterson2014sun}. CUB is a medium-sized fine-grained dataset with 312 attributes, where a total number of 200 classes are presented with a total of 11788 images. CUB is a challenging case as intra-class variance is quite hard to model due to similar appearances and low number of samples. SUN is another medium-sized fine-grained dataset with 102 attributes. SUN, similar to CUB, is a challenging case as it consists of 14340 images of 717 classes, resulting into even fewer images per class compared to CUB. AWA2 is a larger-scale dataset with 85 attributes, where a total of 50 classes are presented with 37322 images. AWA2, although it has a higher amount of images with fewer classes, inherently makes generalization to unseen classes harder. Throughout the experiments, we use the splits proposed in \cite{gbu2018pami} for both ZSL and GZSL experiments. We use the standard per-class top-1 accuracy for ZSL evaluation. For GZSL, per-class top-1 accuracy values for seen and unseen classes are used to compute harmonic-scores. 

\subsection{Implementation Details} \label{imp_det}
In order to make the computational graph end-to-end differentiable, we merge the ResNet-101 \cite{He_2016_CVPR} \textcolor{black}{(used to produce AWA2 \cite{gbu2018pami} dataset embeddings)} feature extractor with ALE model. To reproduce the results of ALE reported in \cite{gbu2018pami}, we freeze the feature extractor and only train ALE for each dataset. In our tables, the reproduced values of ALE are denoted as \textit{original}, although there are slight variations compared to the original results reported by the authors in \cite{gbu2018pami}. We use PyTorch \cite{paszke2019pytorch} for our experiments.

For FGSM, we sweep with a large range of $\epsilon$ values where we end up with visible perturbations. We primarily sweep with \textit{maximum iteration} and \textcolor{black}{$\epsilon$ (added value to cross the boundary)} parameter for DeepFool \textcolor{black}{(DEFO)} and Carlini-Wagner \textcolor{black}{(CaWa, C\&W)} attacks, as we observe diminishing returns \textcolor{black}{(i.e. not producing strong attacks despite reaching intractable compute time)} for other parameters. We assign \textit{0.9} to the ground-truth class in label smoothing defense. For spatial smoothing and total-variance minimization, we use 3x3 windows and maximum iteration of 3, respectively. We apply the same attack and defense parameters for every dataset to facilitate a better comparison of dataset characteristics.
\begingroup
\setlength{\tabcolsep}{8pt} 
\renewcommand{\arraystretch}{1.5} 
\begin{table}[!t]
\resizebox{\textwidth}{!}{%
\begin{tabular}{ccccccccccccc}
\multicolumn{1}{l}{} & \multicolumn{3}{c}{\textbf{Zero Shot}}  & \multicolumn{9}{c}{\textbf{Generalized Zero Shot}}                                                        \\
\multicolumn{1}{l}{} & C    & S    & \multicolumn{1}{c|}{A}    & \multicolumn{3}{c|}{C}                  & \multicolumn{3}{c|}{S}                  & \multicolumn{3}{c}{A} \\ \hline
Attack               & \multicolumn{3}{c|}{Top-1}              & u    & s    & \multicolumn{1}{c|}{h}    & u    & s    & \multicolumn{1}{c|}{h}    & u     & s     & h     \\ \hline
Original             & 54.5 & 57.4 & \multicolumn{1}{c|}{62.0} & 25.6 & 64.6 & \multicolumn{1}{c|}{36.7} & 20.5 & 32.3 & \multicolumn{1}{c|}{25.1} & 15.3  & 78.8  & 25.7  \\ \hline
SpS       & 49.3 & 53.2 & \multicolumn{1}{c|}{59.3} & 21.5 & 56.5 & \multicolumn{1}{c|}{31.1} & 20.1 & 28.0 & \multicolumn{1}{c|}{23.4} & 14.3  & 75.5  & 24.1  \\
LbS         & 52.2 & 55.2 & \multicolumn{1}{c|}{60.6} & 22.7 & 56.2 & \multicolumn{1}{c|}{32.4} & 18.4 & 31.6 & \multicolumn{1}{c|}{23.3} & 16.3  & 74.2  & 26.8  \\
TVM      & 51.4 & 54.0 & \multicolumn{1}{c|}{60.3} & 24.4 & 60.7 & \multicolumn{1}{c|}{34.8} & 19.9 & 29.5 & \multicolumn{1}{c|}{23.8} & 12.9  & 76.4  & 22.1 
\end{tabular}
}
\caption{Results where all images are defended (without any attacks). SpS, LbS and TVM are spatial smoothing, label smoothing and total-variance minimization, respectively.}
\label{tab:defenses}
    \vspace{-8mm}
\end{table}

\endgroup

\subsection{Results} \label{results_pew}

\textbf{Attacks.} First, we present the effect of each attack setting on ZSL/GZSL performance metrics. Results are shown in Table \ref{tab:attacks}. 

In \textit{ZSL} setting, we see every attack has managed to introduce a visible detrimental effect on accuracy values across all datasets. As expected, stronger attacks introduce more pronounced attacks, FGSM being the most effective across all datasets. This is an expected behaviour as we effectively introduce visible and quite strong attacks in the last FGSM setting. In CUB, we see C\&W attack leading in low maximum iterations, but it starts losing out to DeepFool in higher maximum iterations. In SUN, although introducing some effects, C\&W fails to impress and scale with the increasing maximum iteration values, where DeepFool manages to do a better job. In AWA2, C\&W actually does a better job than DeepFool across all parameter settings. FGSM introduces an upward accuracy spike in AWA2, despite its increasing strength. This is primarily caused by actually changing originally incorrectly predicted labels to their correct labels, thereby increasing the accuracy. Lastly, DeepFool produces diminishing returns except the highest maximum iteration setting, across all datasets. 

In \textit{GZSL} setting, we again see an across the board reduction of accuracy values in all datasets. In CUB, DeepFool is the best performing attack, despite FGSM producing significantly more visible perturbations. In SUN,  DeepFool loses out to FGSM slightly, though the produced perturbation is still significantly less visible. For CUB and SUN, DeepFool actually takes about the same time to produce the attack regardless of the maximum iteration value, indicating that it manages to cross the boundary in really few iterations, basically making 10 maximum iterations unnecessary. This means the class boundaries are close to each other and easy to cross, which makes sense as SUN and CUB has significantly more classes compared to AWA2. However, we do not see that effect for C\&W, meaning it still needs more iterations to successfully cross the boundary despite needing the highest compute time. In AWA2, FGSM has a significant lead; DeepFool is somewhat effective but fails to impress. C\&W, on the other hand, basically fails to introduce any meaningful degradation in accuracy, especially in unseen accuracy values. As can be seen from the Table \ref{tab:attacks}, this is actually a wider phenomenon; unseen accuracies are less effected compared to their seen counterparts. We investigate this issue in depth in the following sections.

\begingroup
\setlength{\tabcolsep}{8pt} 
\renewcommand{\arraystretch}{1.5} 
\begin{table}[!t]
\resizebox{\textwidth}{!}{%
\begin{tabular}{cccc|ccccccccc}
\multicolumn{1}{l}{}       & \multicolumn{3}{c|}{\textbf{Zero Shot}} & \multicolumn{9}{c}{\textbf{Generalized Zero Shot}}                                                        \\
\multicolumn{1}{l}{}       & C           & S           & A           & \multicolumn{3}{c|}{C}                  & \multicolumn{3}{c|}{S}                  & \multicolumn{3}{c}{A} \\ \hline
\multicolumn{1}{l}{Attack} & \multicolumn{3}{c|}{Top-1}              & u    & s    & \multicolumn{1}{c|}{h}    & u    & s    & \multicolumn{1}{c|}{h}    & u     & s     & h     \\ \hline
Original                   & 54.5        & 57.4        & 62.0        & 25.6 & 64.6 & \multicolumn{1}{c|}{36.7} & 20.5 & 32.3 & \multicolumn{1}{c|}{25.1} & 15.3  & 78.8  & 25.7  \\ \hline
$FGSM_1$                   & 47.9        & 51.1        & 54.5        & 20.3 & 53.5 & \multicolumn{1}{c|}{29.4} & 19.8 & 26.0 & \multicolumn{1}{c|}{22.5} & 12.7  & 70.0  & 21.5  \\
$FGSM_2$                   & 31.9        & 36.0        & 24.6        & 14.5 & 30.5 & \multicolumn{1}{c|}{19.7} & 14.5 & 16.6 & \multicolumn{1}{c|}{15.5} & 6.2  & 25.3  & 10.0  \\ \hline
$DEFO_1$                   & 46.4        & 49.0        & 58.0        & 18.8 & 50.1 & \multicolumn{1}{c|}{27.3} & 15.9 & 21.0 & \multicolumn{1}{c|}{18.1} & 13.5  & 69.3  & 22.6  \\
$DEFO_3$                   & 46.2        & 48.8        & 58.0        & 18.7 & 50.0 & \multicolumn{1}{c|}{27.2} & 15.9 & 21.0 & \multicolumn{1}{c|}{18.1} & 13.1  & 68.8  & 22.1  \\ \hline
$CaWa_{1}$                 & 48.3        & 52.7        & 58.2        & 21.0 & 55.0 & \multicolumn{1}{c|}{30.5} & 20.2 & 27.3 & \multicolumn{1}{c|}{23.2} & 14.2  & 73.6  & 23.9  \\
$CaWa_{3}$                 & 48.4        & 52.3        & 58.2        & 21.0 & 54.9 & \multicolumn{1}{c|}{30.4} & 20.0 & 27.2 & \multicolumn{1}{c|}{23.1} & 14.2  & 73.3  & 23.8 
\end{tabular}%
}
\caption{Results when \textit{all} images are attacked and then defended with \textit{spatial smoothing}. Parameter sets of the attacks are same as Table \ref{tab:attacks}. }
\label{tab:spatial_smooth}
    \vspace{-8mm}
\end{table}
\endgroup

\textbf{Defenses.} Before going through the recovery rates of each defense, we first apply the defenses \textit{without} any attacks to see what the effects of defenses are; a defense that is actually degrading the results are naturally not suitable for use. Results are shown in Table \ref{tab:defenses}. We see modest detrimental effects of defenses across the board, which we believe to be acceptable given the improvements they bring. We also see that in AWA2, label smoothing actually improves the GZSL performance compared to its original value. There is no winner in this regard, although label smoothing and total-variance minimization tend to do a better job than spatial smoothing. We now analyze the effects of each defense under several attack settings; we note that we omit one setting per each attack algorithm from our defense analysis; they either introduce extreme perturbations ($FGSM_3$) or negligible effects compared to their weaker counterpart ($DeepFool_2$ and $C\&W_2$)

\textit{Spatial smoothing} results are shown in Table \ref{tab:spatial_smooth} \footnote{Tables \ref{tab:spatial_smooth}, \ref{tab:label_smooth} and \ref{tab:tvm_defense} should be compared to Table \ref{tab:attacks}.}.  In \textit{ZSL} setting, we see quite good recoveries across all datasets. The recovered accuracy values are naturally better for weaker attacks. We see quite similar recovered accuracy values for each DeepFool and C\&W settings ($DeepFool_1$ vs $DeepFool_3$, $C\&W_1$ vs $C\&W_3$), in constrast with what we see for FGSM. This is potentially due to the nature of the attacks; FGSM strength scales proportionally with the coefficient $\epsilon$, whereas maximum iteration for C\&W and DeepFool acts like a binary switch indicating whether the attacks will function or not. In \textit{GZSL} setting, results generally show the same trends with ZSL. However, we see negligible recoveries for C\&W and DeepFool in AWA2, especially in unseen accuracy values. Surprisingly, spatial smoothing \textit{degrades} the unseen and harmonic scores of $C\&W_1$  even compared to its original (unattacked) values. This phenomenon will be investigated in-depth in the following sections.

\textit{Label smoothing} results are shown in Table \ref{tab:label_smooth}. Although we can not directly compare recovered accuracies to ones reported in Table \ref{tab:attacks} as the original accuracy is different, we compare the trends to gain insights for label smoothing. In \textit{ZSL} setting,  we see no visible gains for FGSM and even reductions in accuracy for some cases.  DeepFool results are actually improved for some cases and C\&W sees the most dramatic improvements among all attacks. In \textit{GZSL}, similar trends with ZSL is observed; FGSM seems unaffected, DeepFool is slightly recovered but C\&W is the most recovered attack.

\textit{Total-variance minimization} results are shown in Table \ref{tab:tvm_defense}. In \textit{ZSL} setting, we observe across the board recoveries for every attack setting. Similar to what we observed in \textit{spatial smoothing}, recovered values for DeepFool and C\&W are similar in values. Compared to other defenses, TVM does the best job in ZSL accuracy recovery. In \textit{GZSL} setting, we see similar trends with ZSL. However, we observe in AWA2 that unseen accuracies actually go down when TVM is applied, especially for DeepFool and C\&W. For C\&W, this effect is also present for harmonic scores. This is similar to what we observed in spatial smoothing, however the effect is more pronounced. This will be investigated later in the paper.

\begin{figure}[!t]
\begin{center}
      \includegraphics[width=1.0\textwidth]{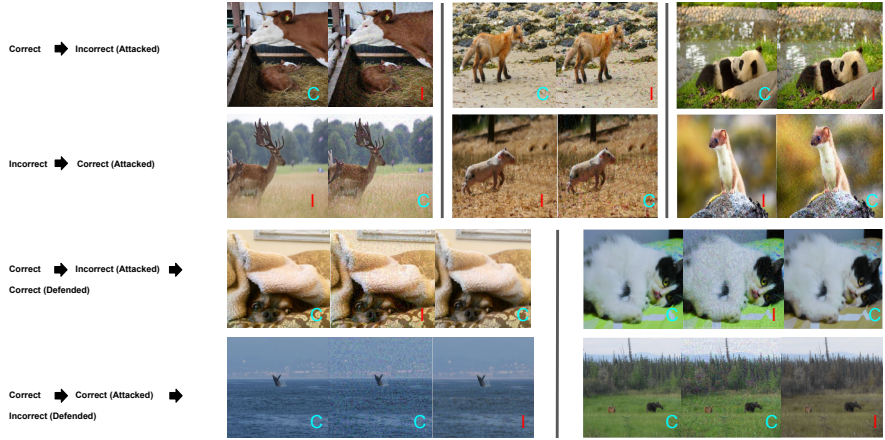}
        \vspace{-10mm}
    \caption{Example images from AWA2 dataset. The first and second rows show (in pairs) original and perturbed images, where attacks have induced misclassification and converted mispredictions into correct predictions, respectively. Third and fourth rows show (in triples) original, attacked and defended images where defenses have corrected and induced misclassifications, respectively. We use a powerful attack to show a more visible perturbation. I and C indicate incorrect and correct predictions.}
    \label{fig:qualitativeResults}
    \vspace{-8mm}
\end{center}
\end{figure}

\textbf{Summary.} In \textit{attacks}, an unbounded, high epsilon FGSM attack is the strongest and the fastest one, as expected. However, when minimum perturbation is considered, FGSM loses out to DeepFool and C\&W significantly. Across all datasets, DeepFool seems to be the best trade-off between perturbation magnitude and attack success. In \textit{defenses}, we see varying degrees of success for each dataset. In CUB, we see spatial smoothing to the best for FGSM attacks, whereas TVM is the best for the rest. In AWA2, spatial smoothing is the best all-around defense for every attack setting. For SUN, spatial smoothing is still the best for FGSM, however TVM has a lead in C\&W and DeepFool. Label smoothing is the worst defense all around and TVM is the most compute-heavy one, as expected. We present qualitative samples in Figure \ref{fig:qualitativeResults}.

\begingroup
\setlength{\tabcolsep}{8pt} 
\renewcommand{\arraystretch}{1.5} 
\begin{table}[!t]
\resizebox{\textwidth}{!}{%
\begin{tabular}{cccc|ccccccccc}
\multicolumn{1}{l}{}       & \multicolumn{3}{c|}{\textbf{Zero Shot}} & \multicolumn{9}{c}{\textbf{Generalized Zero Shot}}                                                        \\
\multicolumn{1}{l}{}       & C           & S           & A           & \multicolumn{3}{c|}{C}                  & \multicolumn{3}{c|}{S}                  & \multicolumn{3}{c}{A} \\ \hline
\multicolumn{1}{l}{Attack} & \multicolumn{3}{c|}{Top-1}              & u    & s    & \multicolumn{1}{c|}{h}    & u    & s    & \multicolumn{1}{c|}{h}    & u      & s     & h    \\ \hline
Original                   & 52.2        & 55.2        & 60.6        & 22.7 & 56.2 & \multicolumn{1}{c|}{32.4} & 18.4 & 31.6 & \multicolumn{1}{c|}{23.3} & 16.3   & 74.2  & 26.8 \\ \hline
$FGSM_1$                   & 39.8        & 46.8        & 41.1        & 17.4 & 43.7 & \multicolumn{1}{c|}{24.9} & 15.7 & 25.2 & \multicolumn{1}{c|}{19.4} & 12.1   & 59.8  & 20.1 \\
$FGSM_2$                   & 11.7        & 15.2        & 14.1        & 6.7 & 9.8 & \multicolumn{1}{c|}{0.80} & 5.5 & 7.7 & \multicolumn{1}{c|}{6.4} & 2.7   & 10.3  & 4.4 \\ \hline
$DEFO_1$                   & 29.8        & 30.4        & 49.6        & 10.0 & 20.3 & \multicolumn{1}{c|}{13.4} & 6.2 & 8.8 & \multicolumn{1}{c|}{7.3} & 14.01  & 42.4  & 21.1 \\
$DEFO_3$                   & 19.8        & 19.2        & 41.7        & 8.2 & 11.9 & \multicolumn{1}{c|}{9.7} & 5.2 & 7.1 & \multicolumn{1}{c|}{6.0} & 13.1   & 25.5  & 17.3 \\ \hline
$CaWa_{1}$                 & 38.8        & 45.6        & 46.6        & 19.4 & 40.4 & \multicolumn{1}{c|}{26.3} & 16.2 & 26.6 & \multicolumn{1}{c|}{20.1} & 16.4   & 61.0  & 25.9 \\
$CaWa_{3}$                 & 34.1        & 42.7        & 40.6        & 18.8 & 34.8 & \multicolumn{1}{c|}{24.5} & 16.2 & 25.1 & \multicolumn{1}{c|}{19.7} & 16.2   & 52.0  & 24.7
\end{tabular}%
}
\caption{Results when \textit{all} images are attacked and then defended with \textit{label smoothing}. Parameter sets of the attacks are same as Table \ref{tab:attacks}. \textit{Original} results are results obtained by training ALE with label-smoothing.}
\label{tab:label_smooth}
    \vspace{-8mm}
\end{table}
\endgroup

\subsection{Analysis}

\begingroup
\setlength{\tabcolsep}{8pt} 
\renewcommand{\arraystretch}{1.5} 
\begin{table}[!t]
\resizebox{\textwidth}{!}{%
\begin{tabular}{cccc|ccccccccc}
\multicolumn{1}{l}{} & \multicolumn{3}{c}{\textbf{Zero Shot}}  & \multicolumn{9}{c}{\textbf{Generalized Zero Shot}}                                                        \\
\multicolumn{1}{l}{} & C    & S    & \multicolumn{1}{c|}{A}    & \multicolumn{3}{c|}{C}                  & \multicolumn{3}{c|}{S}                  & \multicolumn{3}{c}{A} \\ \hline
Attack               & \multicolumn{3}{c|}{Top-1}              & u    & s    & \multicolumn{1}{c|}{h}    & u    & s    & \multicolumn{1}{c|}{h}    & u     & s     & h     \\ \hline
Original             & 54.5 & 57.4 & \multicolumn{1}{c|}{62.0} & 25.6 & 64.6 & \multicolumn{1}{c|}{36.7} & 20.5 & 32.3 & \multicolumn{1}{c|}{25.1} & 15.3  & 78.8  & 25.7  \\ \hline
$FGSM_1$             & 49.1 & 53.2 & \multicolumn{1}{c|}{53.8} & 23.0 & 57.5 & \multicolumn{1}{c|}{32.8} & 18.9 & 28.3 & \multicolumn{1}{c|}{22.7} & 11.7  & 71.8  & 20.1  \\
$FGSM_2$             & 25.3 & 32.8 & \multicolumn{1}{c|}{21.1} & 12.6 & 21.9 & \multicolumn{1}{c|}{16.0} & 12.6 & 15.3 & \multicolumn{1}{c|}{13.8} & 5.0  & 22.5  & 8.2  \\ \hline
$DEFO_1$             & 48.4 & 50.3 & \multicolumn{1}{c|}{59.0} & 19.7 & 52.3 & \multicolumn{1}{c|}{28.6} & 15.2 & 20.8 & \multicolumn{1}{c|}{17.5} & 12.5  & 70.9  & 21.4  \\
$DEFO_3$             & 48.3 & 50.3 & \multicolumn{1}{c|}{59.0} & 19.5 & 52.3 & \multicolumn{1}{c|}{28.4} & 15.1 & 20.8 & \multicolumn{1}{c|}{17.5} & 12.5  & 70.6  & 21.3  \\ \hline
$CaWa_{1}$           & 50.9 & 53.3 & \multicolumn{1}{c|}{58.8} & 24.0 & 60.3 & \multicolumn{1}{c|}{34.3} & 20.0 & 29.2 & \multicolumn{1}{c|}{23.8} & 12.7  & 75.6  & 21.7  \\
$CaWa_{3}$           & 51.2 & 53.4 & \multicolumn{1}{c|}{58.9} & 24.2 & 60.2 & \multicolumn{1}{c|}{34.4} & 19.9 & 29.1 & \multicolumn{1}{c|}{23.6} & 12.6  & 75.6  & 21.6  
\end{tabular}%
}
\caption{Results when \textit{all} images are attacked and then defended with \textit{total-variance minimization}. Parameter sets of the attacks are same as in Table \ref{tab:attacks}.}
\label{tab:tvm_defense}
    \vspace{-8mm}
\end{table}
\endgroup

It is clear that adversarial examples can be considered as out-of-distribution samples which we fail to recognize properly. As they do not have their own class prototypes, the learned ranking system incorrectly assigns them to a class. Effectively, we require a mechanism to transfer knowledge from clean to adversarial images, on top of the seen-to-unseen transfer we need to tackle already. Moreover, possibly from a simpler perspective, ZSL models can be considered as immature compared to supervised models; accuracy levels are not on the same level. The second perspective harbors interesting facts. Assuming a model with the perfect accuracy, we know attacks can only degrade the results, assuming they are effective. Defenses can still degrade the results without any attacks, but we know they alleviate the issues to a certain degree, assuming they are effective. What happens when the model is far from perfect is what we focus on now. 

\textbf{Class-transitions: False/Correct.} It is observed throughout the attacks that in \textit{GZSL} setting, unseen accuracies are less effected compared to seen accuracies. We further investigate this by looking at the class-transitions during each attack setting. For each class, we calculate the ratio of class transitions; out of all (originally) correctly predicted samples, what percentage have transitioned to false? Out of all (originally) falsely predicted samples, what percentage have transitioned to correct or \textit{other} false classes? Our results are shown in Table \ref{tab:correct_false_transitions}.

Stronger the attack, higher \textcolor{black}{\textit{correct-to-false}} (CF) percentages we observe. Moreover, stronger attacks also introduce higher \textcolor{black}{\textit{false-to-other-false}} (FF) ratios. This simply means regardless of the success of original predictions, stronger attacks induce more class transitions. Statistically, there is also the possibility of an attack \textit{correcting} an originally incorrect prediction. We observe the highest FC ratios in C\&W attacks and the lowest in DeepFool attacks. Coupled with the lowest CF ratios, this can be the reason as to why C\&W performed the worst in our attack scenarios. 

\begingroup
\setlength{\tabcolsep}{6pt} 
\renewcommand{\arraystretch}{1.5} 
\begin{table}[]
\resizebox{\textwidth}{!}{%
\begin{tabular}{cclcclccccllcclllll}
\multicolumn{1}{l}{} & \multicolumn{18}{c}{\textbf{Generalized Zero Shot}}                                                                                                                                                                                                                                                                               \\
\multicolumn{1}{l}{} & \multicolumn{6}{c|}{C}                                                                                        & \multicolumn{6}{c|}{S}                                                                                                            & \multicolumn{6}{c}{A}                                                         \\ \hline
Class Type           & \multicolumn{3}{c|}{U}                                & \multicolumn{3}{c|}{S}                                & \multicolumn{3}{c|}{U}                                                    & \multicolumn{3}{c|}{S}                                & \multicolumn{3}{c|}{U}                                & \multicolumn{3}{c}{S} \\ \hline
Transitions          & \multicolumn{1}{l}{CF} & FC & \multicolumn{1}{l|}{FF} & \multicolumn{1}{l}{CF} & FC & \multicolumn{1}{l|}{FF} & \multicolumn{1}{l}{CF} & \multicolumn{1}{l}{FC} & \multicolumn{1}{l|}{FF} & CF                     & FC & \multicolumn{1}{l|}{FF} & \multicolumn{1}{l}{CF} & FC & \multicolumn{1}{l|}{FF} & CF    & FC    & FF    \\ \hline
$FGSM_1$             & 81                     & 20 & \multicolumn{1}{c|}{69} & 55                     & 40 & \multicolumn{1}{c|}{47} & 65                     & 11                     & \multicolumn{1}{c|}{65} & \multicolumn{1}{c}{54} & 16 & \multicolumn{1}{c|}{59} & 89                     & 12 & \multicolumn{1}{l|}{69} & 39    & 49    & 43    \\
$FGSM_2$             & 99                     & 18 & \multicolumn{1}{c|}{82} & 99                     & 36 & \multicolumn{1}{c|}{64} & 99                     & 12                     & \multicolumn{1}{c|}{87} & 98                     & 16 & \multicolumn{1}{c|}{84} & 100                    & 5  & \multicolumn{1}{l|}{93} & 94    & 33    & 67    \\ \hline
$DEFO_1$             & 91                     & 10 & \multicolumn{1}{c|}{82} & 83                     & 21 & \multicolumn{1}{c|}{70} & 93                     & 8                      & \multicolumn{1}{c|}{89} & 95                     & 10 & \multicolumn{1}{c|}{87} & 53                     & 7  & \multicolumn{1}{l|}{61} & 55    & 26    & 44    \\
$DEFO_3$             & 94                     & 11 & \multicolumn{1}{c|}{84} & 93                     & 22 & \multicolumn{1}{c|}{72} & 94                     & 8                      & \multicolumn{1}{c|}{90} & 98                     & 10 & \multicolumn{1}{c|}{88} & 60                     & 7  & \multicolumn{1}{l|}{72} & 74    & 28    & 49    \\ \hline
$CaWa_{1}$           & 92                     & 24 & \multicolumn{1}{c|}{69} & 78                     & 39 & \multicolumn{1}{c|}{51} & 63                     & 12                     & \multicolumn{1}{c|}{61} & 55                     & 15 & \multicolumn{1}{c|}{57} & 81                     & 16 & \multicolumn{1}{l|}{58} & 43    & 53    & 36    \\
$CaWa_{3}$           & 95                     & 24 & \multicolumn{1}{c|}{70} & 86                     & 40 & \multicolumn{1}{c|}{53} & 67                     & 13                     & \multicolumn{1}{c|}{66} & 61                     & 10 & \multicolumn{1}{c|}{90} & 89                     & 18 & \multicolumn{1}{l|}{64} & 60    & 57    & 38   
\end{tabular}%
}
\caption{Categorization of prediction changes induced by each attack. U and S columns are results for unseen and seen classes, respectively. CF, FC and FF are \textit{correct-to-false } (as the percentage of all originally correct predictions), \textit{false-to-correct} and \textit{false-to-other-false} (as the percentage of all originally incorrect predictions) changes in \%, represented as per-class normalized ratio averages. Classes having no originally correct or incorrect predictions have not been included in the calculation.}
\label{tab:correct_false_transitions}
    \vspace{-8mm}
\end{table}
\endgroup

When we compare seen and unseen classes, we see higher FC ratios for seen classes. Moreover, seen classes have smaller CF ratios which means seen classes are less effected detrimentally and more effected positively. This contradicts with our starting point; unseen classes being less effected by an attack than seen classes. However, unseen classes have a lot of initially zero accuracy classes and they are not taken into account in our calculation. This leads to fewer number of classes with higher than 0 accuracy and fewer correctly predicted samples for each (as unseen accuracies are low all around). Once these samples are effected, we observe higher FC rates. However, high FC rates in seen classes tells us that in an event of misclassification, the algorithm predicts the correct class with a high probability, but not high enough to be the highest prediction. This is an interesting effect of the adversarial attacks, which means softmax probabilities are quite close to each other, and class boundary transition is easier. Conversely, one can expect high CF ratios for seen classes, but this is not the case. However, it can be related to the fact that the model, for seen classes, is robust against attacks when it comes to correct predictions, but its false predictions are not really confident.

\textbf{Class-transitions: Seen/Unseen.} We also analyze the effect of attacks from a seen/unseen class perspective. For each class, we calculate the following for each class and average it for seen and unseen classes; out of \textit{all} changed samples, what percent went to a seen or an unseen class? Our results are shown in Table \ref{tab:class_transitions}. Results show that except FGSM, attack behaviour in terms of seen/unseen class transition seems to be stable. For FGSM, however, we see increase in unseen-to-seen transitions, which is in line with the further decrease of accuracy values (i.e. unseen-to-unseen can have false to correct transitions for unseen). This behaviour bodes well with the attack settings; FGSM scales its attack with the $\epsilon$ coefficient whereas DeepFool and C\&W simply have more time to \textit{solve} for the minimum perturbation with higher maximum iterations. Regardless of the dataset and the attack setting, an overwhelming majority of the transitions happen towards seen classes. This is likely due to the fact that the model trains exclusively on seen classes and naturally is more confident about its predictions, and this can cause a bias towards seen classes in the class boundary transitions.

\begingroup
\setlength{\tabcolsep}{8pt} 
\renewcommand{\arraystretch}{1.5} 
\begin{table}[!t]
\resizebox{\textwidth}{!}{%
\begin{tabular}{cccclcccclccl}
\multicolumn{1}{l}{} & \multicolumn{12}{c}{\textbf{Generalized Zero Shot}}                                                                                                                                                                                                                                     \\
\multicolumn{1}{l}{} & \multicolumn{4}{c|}{C}                                                                             & \multicolumn{4}{c|}{S}                                                                             & \multicolumn{4}{c}{A}                                                         \\ \hline
Class Transition     & \multicolumn{1}{l}{UU} & \multicolumn{1}{l}{US} & \multicolumn{1}{l}{SU} & \multicolumn{1}{l|}{SS} & \multicolumn{1}{l}{UU} & \multicolumn{1}{l}{US} & \multicolumn{1}{l}{SU} & \multicolumn{1}{l|}{SS} & UU                     & \multicolumn{1}{l}{US} & \multicolumn{1}{l}{SU} & SS \\ \hline
$FGSM_1$             & 30                     & 70                     & 16                     & \multicolumn{1}{l|}{84} & 22                     & 78                     & 10                     & \multicolumn{1}{c|}{90} & \multicolumn{1}{c}{17} & 83                     & 7                     & 93 \\
$FGSM_2$             & 28                     & 72                     & 18                     & \multicolumn{1}{l|}{82} &17                     & 83                     & 10                     & \multicolumn{1}{c|}{90} & 12                     & 88                     & 7                     & 93 \\ \hline
$DEFO_1$             & 24                     & 76                     & 20                     & \multicolumn{1}{l|}{80} & 16                     & 84                     & 10                     & \multicolumn{1}{c|}{90} & 13                     & 87                     & 7                     & 93 \\
$DEFO_3$             & 24                     & 76                     & 20                     & \multicolumn{1}{l|}{80} & 16                     & 84                     & 10                     & \multicolumn{1}{c|}{90} & 14                     & 86                     & 7                     & 93 \\ \hline
$CaWa_{1}$           & 31                     & 69                     & 17                     & \multicolumn{1}{l|}{83} & 22                     & 78                     & 10                     & \multicolumn{1}{c|}{90} & 24                     & 76                     & 8                     & 92 \\
$CaWa_{3}$           & 31                     & 69                     & 17                     & \multicolumn{1}{l|}{83} & 22                     & 78                     & 10                     & \multicolumn{1}{c|}{90} & 25                     & 75                     & 9                     & 91
\end{tabular}%
}
\caption{Attack-induced, per-class normalized class transition averages (in \%) for different attack settings. UU, US, SU and SS are unseen-to-unseen, unseen-to-seen, seen-to-unseen and seen-to-seen transitions, respectively.}
\label{tab:class_transitions}
    \vspace{-8mm}
\end{table}
\endgroup
\textbf{Adverse effects of defenses.} As observed in Section \ref{results_pew}, there have been cases where defenses actually reduced the accuracy after the attacks rather than recovering it. Following the work we've presented in Table \ref{tab:correct_false_transitions}, we observe the effect of defenses (i.e. we add another layer to CF, FC, FF transitions, such as CFC, FCF, FFC, etc) \footnote{We do not include a table for this analysis due to space restrictions.}. Logically, we can analyse the effect of defenses in four main categories; it corrects a mistake (CFC, FFC), it preserves the results (CCC, FFF) and it has detrimental effects (CCF, FCF) and it fails to recover (CFF, FCC). It must be noted that recovery here means recovering the \textit{original} label, not necessarily the correct one. Across all experiments we observe every category of effect up to some degree, with correct-recoveries (CFC) spearheading the overall recovery of accuracy values. However, we observe that the defense-induced reduction of accuracies strongly correlate with high FCF ratios. This means alleviating the \textit{positive} effects of attacks. Although the defense does its job by recovering the original predictions, the reductions occur nonetheless.

\textbf{Attacking only correct predictions.}  We investigate attacking only the originally correct predictions and only defending them. This is not under the threat model we assumed in the beginning, but it is valuable to decouple weak model effects (i.e. low accuracy) with potential ZSL-specific effects in our results. We do not include numerical results due to space restrictions.

We naturally observe that the \textit{unintuitive} effects such as attacks \textit{improving} the results or defences \textit{degrading} the results eliminated. Across all attacks, we see more dramatic accuracy reductions and we see improvements across all defenses. The overall \textit{rankings} for best attack and defense follow our previous \textit{all images} attack settings. In this setting, results are more reminiscent of a supervised model, however due to the extreme bias between seen and unseen classes, GZSL-specific effects still remain (i.e. unseen and seen classes being effected differently, class transition trends).

\textbf{Dataset characteristics.} The datasets considered are inherently different; SUN and CUB have fewer samples per class and consist of high number of classes, whereas AWA2 has high number of samples per class but consist of fewer classes. In AWA2, we see attacks failing to effect in their weakest setting; especially DeepFool and C\&W performing their worst among other datasets. This is likely related to the sample count of AWA2; a larger distribution per class helps robustness, as suggested in \cite{schmidt2018adversarially}. We see FC transitions happening more frequently compared to other datasets; this is likely an effect of multiple confident predictions as this effect is more pronounced in seen classes. Upward accuracy spikes that occur in FGSM attacks (this analysis is performed with a wide range of parameters for FGSM and not included in detail due to page limitations) are more frequent here as well (especially in ZSL setting); this is likely an \textcolor{black}{effect} of having fewer number of classes as misclassifications are statistically more likely to fall into the originally correct classes. We see similar trends for SUN and CUB in general; SUN has the fewest transitions to unseen classes. This correlates strongly with the really high number of classes in SUN. In overall, we see SUN and CUB get better returns from all defenses, compared to AWA2.

%% file: conclusion.tex
\section{Conclusion and Future Work} \label{conclude}
Despite their stunning success, it is shown that machine learning models can be fooled with carefully crafted perturbations. Adversarial robustness have generally been studied from a fully supervised perspective. ZSL and GZSL algorithms that lack supervision for a set of classes have not received attention for their adversarial robustness. In this paper, we introduce a study aiming to fill this gap by assessing a well-known ZSL model for its adversarial robustness, both ZSL and GZSL evaluation set-ups. We subject the model to several attacks and defenses across widely-used ZSL datasets. Our results indicate that adversarial robustness for ZSL has its own challenges, such as the extreme data bias and the comparably immature state of the field (compared to supervised learning). We highlight and analyse several points, especially in GZSL settings, to guide future researchers as to what needs attention in making ZSL models robust and also what points could be important for interpreting the results.